\documentclass[letterpaper]{article}
\usepackage{aaai18}
\usepackage{times}
\usepackage{helvet}
\usepackage{courier}
\usepackage{graphicx}
\usepackage{amsmath}
\usepackage{hyperref}
\usepackage{multirow}
\usepackage[ruled,norelsize]{algorithm2e}
\begin{document}
\title{MEETING BOT: Reinforcement Learning for Dialogue Based\\Meeting Scheduling}
\author{Vishwanath D,\quad Lovekesh Vig,\quad Gautam Shroff \enspace \& \enspace Puneet Agarwal\\
\Large{TCS Research, New Delhi}\\
\texttt{\{vishwanath.d2, lovekesh.vig, gautam.shroff, puneet.a\}@tcs.com}
}

\maketitle
\begin{abstract}
In this paper we present Meeting Bot, a reinforcement learning based conversational system that interacts with multiple users to schedule meetings. The system is able to interpret user utterences and map them to preferred time slots, which are then fed to a reinforcement learning (RL) system with the goal of converging on an agreeable time slot. The RL system is able to adapt to user preferences and environmental changes in meeting arrival rate while still scheduling effectively. Learning is performed via policy gradient with exploration, by utilizing an MLP as an approximator of the policy function. Results demonstrate that the system outperforms standard scheduling algorithms in terms of overall scheduling efficiency. Additionally, the system is able to adapt its strategy to situations when users consistently reject or accept meetings in certain slots (such as Friday afternoon versus Thursday morning), or when the meeting is called by members who are at a more senior designation.
\end{abstract}

\section{Introduction}
One of the most frequently performed tasks in an organization is the scheduling of meetings between employees across different designations and timezones. This scheduling of meetings is frequently performed over email, or by human assistants, often involving several back and forth negotiations over the actual time slot of the meeting. In this paper we present a learning system whereby a user is able to converse with a virtual assistant to convey his/her desire to initiate a meeting with a set of participants at some preferred range of timeslots. The virtual assistant then interacts with the meeting participants (including the initiator if necessary) via dialogue to converge on an agreeable timeslot. The system attempts to schedule the meeting while trying to optimize two different objectives; the first objective being to schedule all the meetings in the system efficiently i.e.  maximizing the number of meetings scheduled, and the second being to minimize the number of interactions with the meeting participants, to avoid annoying them or wasting their time. The meeting is considered scheduled when all participants agree on a slot(s).   

There are several natural language and learning components to such a scheduling system. Initially the virtual assistant has to extract the initiators desired timeslots from his/her utterances, in addition to the names of the meeting participants. While the names of the participants can be extracted using a standard Named Entity Recognition (NER) engine like Stanford NER \cite{StanNER}, the extraction of the correct timeslots can be extremely challenging as the initiator may express the desired time via vague natural language utterances such as 'please schedule a meeting for Thursday morning'  or 'Friday or Monday is preferred but avoid scheduling in the morning'. We propose a multi-label learning approach to map initiator utterances to multiple possible time slots. The approach utilizes a Long Short Term Memory (LSTM) \cite{hochreiter1997long} based multi-label learning \cite{tsoumakas2006multi} model that utilizes independent loss functions over the output units and successfully learns to map initiator utterances to correct timeslots with high accuracy. 

Once the desired set of timeslots is determined, the virtual assistant employs policy gradient based reinforcement learning \cite{sutton1998introduction} to decide on the correct slots to schedule the meeting. The reinforcement learning agent is aware of all the meetings waiting to be scheduled in the system and the current availability of different users in the system for all the slots over the coming week. Based on the meeting traffic and the prior experience of scheduling meetings with different users in the system, the agent attempts to choose slots in which it estimates the maximum probability of the participants agreeing. The  state space for the reinforcement learner comprises the information about 1) the current occupancy of the slots, 2) the meetings waiting in the queue, 3) the initiator IDs of all the meetings, and 4) the designations of all the participants of the meeting. The environment automatically decides upon a meeting and a slot based on a bin packing scheduling heuristic. At each step, the agent has to decide between two possible actions i.e. whether or not to attempt to schedule the selected meeting at the chosen slot by issuing a request to the meeting participants. If the agent decides to schedule, a dialogue is initiated wherein the agent must request the participants to attend the meeting at the chosen slot. The reply is processed to determine whether the user agreed and the resulting experiences are rewarded with both immediate and delayed rewards to balance the tradeoff between scheduling effectively avoiding unnecessary meeting requests.

Experiments are performed to demonstrate the scheduling efficiency of the system and the ability of the system to modify its policy based on differing meeting arrival rates. Additionally, we examine the adaptability of the system to user preferences for certain slots, and to the designations of the meeting initiators. Results are promising and demonstrate that the system learns a robust adaptive policy for scheduling meetings via dialogue while minimizing the number of meeting requests to users. 

The primary contributions of this paper are as follows:
\begin{enumerate}
 \item A recurrent multi label classification model to map natural language user utterances to desired time slots.
 \item A policy gradient based reinforcement learning framework with global reward to learn a versatile adaptive policy for scheduling meetings.
 \item The use of both immediate and delayed rewards to learn user based preferences for time slots and initiator designations.
 \item To the best of our knowledge, this is the first attempt at utilizing reinforcement learning to schedule meetings based on user preferences.
\end{enumerate}

\section{Related Work}
This paper draws upon several ideas from work in planning and scheduling, deep learning for natural language understanding, and reinforcement learning. The use of recurrent neural models such as Long Short Term Memory (LSTM) \cite{hochreiter1997long} networks and Gated Recurrent Units (GRUs) \cite{chung2014empirical} to model text sequences has become commonplace. However, there has been limited utilization of these networks for multi-label classification tasks \cite{lipton2015learning}, \cite{nigam16} which is one of the focal problems in this paper.

\subsection{Reinforcement Learning for Scheduling}
Reinforcement learning (RL) approaches involve an agent in an environment learning a policy for choosing appropriate actions in different states so as to maximize the agents environmental reward. More formally, given a state $s_t$, the policy determines which action $a_t$ the agent should perform to transition to state $s_{t+1}$ and receive reward $r_t$ from the environment. The policy is defined as a probability distribution over the actions $\pi(s,a)$ and it is hard to learn the correct action for each state by experience simply because the state space for most real world problems is exponentially large. Hence, function approximators such as deep neural networks have been used for approximating the policy function, and allow for generalization of the learnt policy function across states. Thus, the problem reduces to finding the correct  parameters $\theta$ for the chosen policy function approximator (in this case a deep feedforward neural network).

Despite the recent promise shown by deep reinforcement learners in complex gaming environments like Go \cite{silver2016mastering} and for increasing energy efficiency of cooling systems \cite{cooling}, the use of RL for scheduling problems has been limited\cite{zhang1995reinforcement}, \cite{mao2016resource}. We contend that utilizing reinforcement learning has advantages over traditional heuristic based solutions to such problems because:

\begin{enumerate}
 \item RL based learners can utilize knowledge from past schedules and learn patterns that lead to good solutions
 \item RL based learners can adapt to changing conditions in the scheduling environment such as user preferences, novel constraints.
\end{enumerate}

In this paper we utilize the method of policy gradients to learn a policy function that maximizes future reward. The idea is that we attempt to move the policy towards state-actions in proportion to the expected cumulative reward obtained by following the policy thereafter. Assuming that future rewards are discounted by a discount factor $\gamma$,  the net future discounted reward from time $t$ is given by $\sum_{t=0}^{\infty}\gamma^{t}r_t$. The policy objective is to maximize the expected cumulative discounted reward. To achieve this policy gradient methods evaluate the gradient of the policy objective given by \cite{williams1992simple}, \cite{sutton1998introduction}:
\begin{equation}
 \nabla_{\theta}E_{\pi_{\theta}}[\sum_{t=0}^{\infty}\gamma^{t}r_t]=E_{\pi_{\theta}}[\nabla \log\pi_{\theta}(s,a)Q^{\pi_{\theta}}(s,a)]
\end{equation}

In order to evaluate the gradient a popular approach is to sample the reward obtained from several trajectories or episodes where the agent follows the given policy, and use the  empirical cumulative discounted reward $r_t$ as an unbiased estimator for the value function $Q$. Gradient descent then yields the updated policy parameters:

\begin{equation}
 \theta=\theta + \alpha\sum_t \nabla_\theta \log\pi_{\theta}(s_t , a_t)r_t
\end{equation}

\section{System Architecture}
There are two dialogue based interactions that take place in our system:
\begin{itemize}
 \item A dialogue with the initiator to obtain the details about the meeting, such as the names/ids of the participants and the proposed meeting time. 
 \item An RL based dialogue with the different participants to request their availability for the time slot in question.
\end{itemize}

\begin{figure}[ht]
\includegraphics[width=\columnwidth]{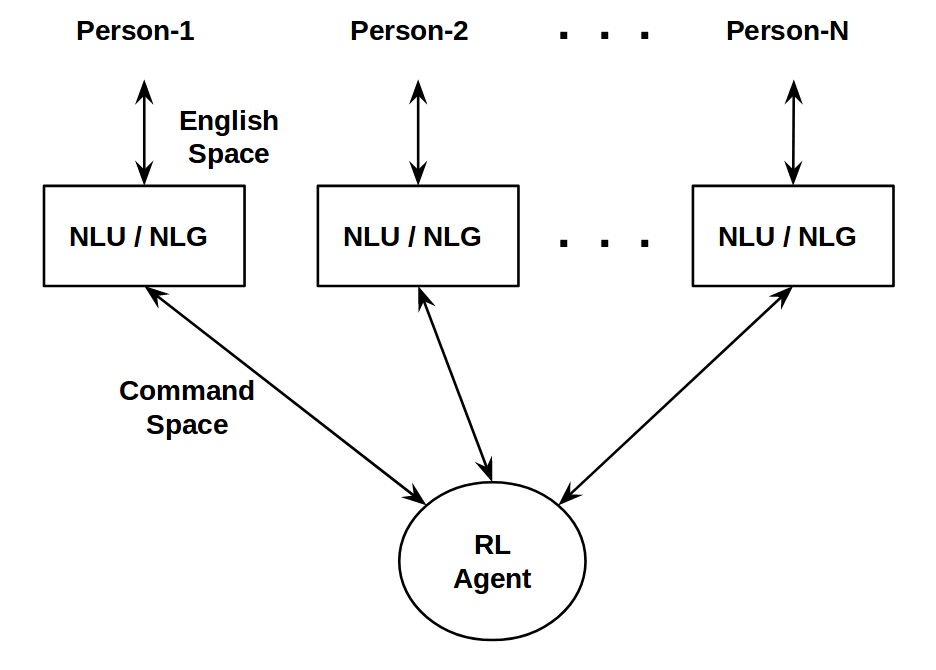}
\caption{System Architecture}
\label{fig:big_picture}
\end{figure}

\subsection{Dialogue for Meeting Initiation}
\begin{figure}[ht]
\includegraphics[width=\columnwidth]{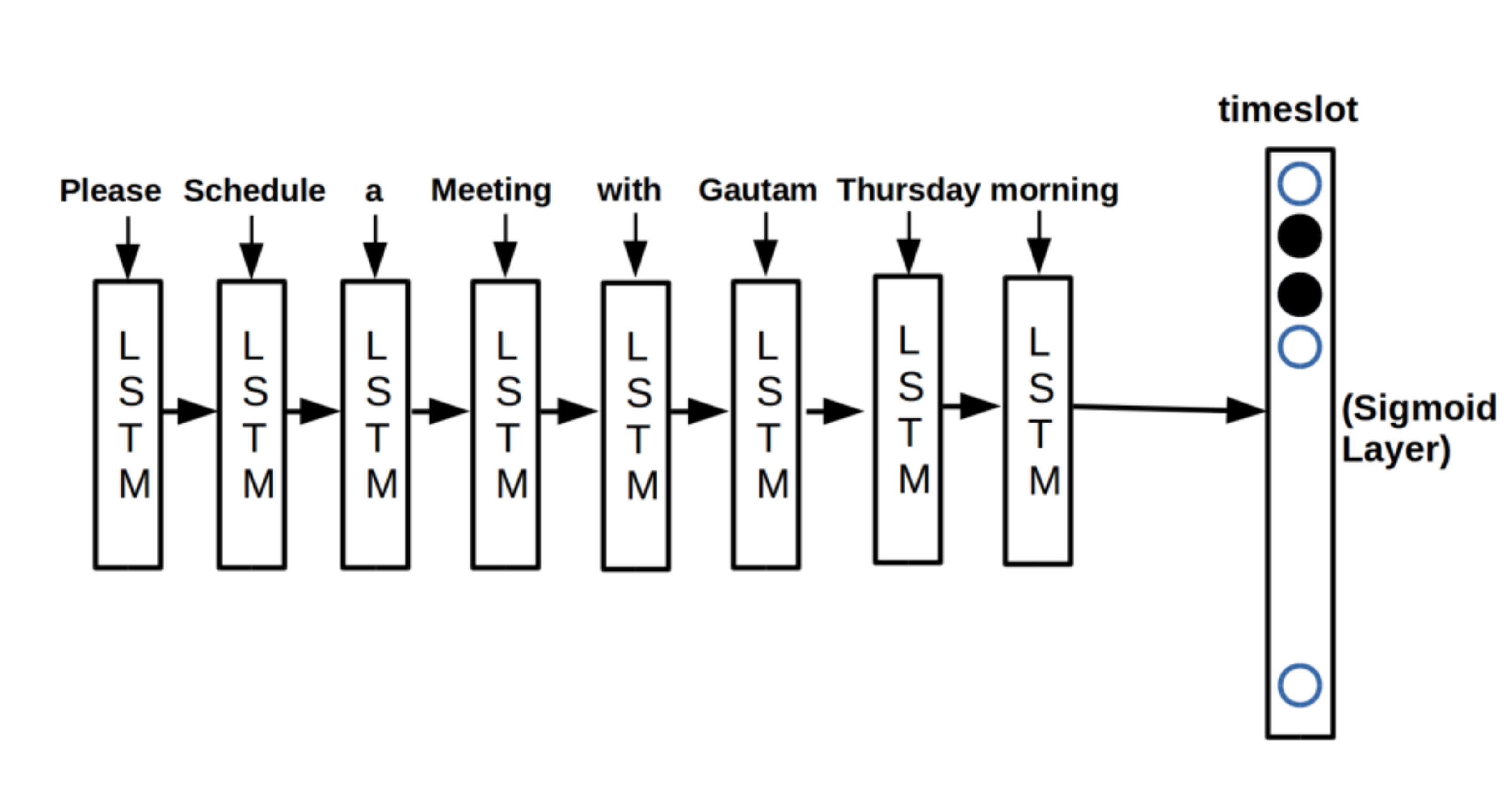}
\caption{LSTM Architecture for converting sentence which has time information into slots}
\label{fig:slotmap}
\end{figure}

Here the sentence entered by the user requesting a meeting is fed into an LSTM model that is trained to map sentences to appropriate slots. For instance the sentence "Please schedule a meeting with Gautam for Wednesday afternoon`` is mapped to the slots corresponding to Wednesday afternoon. Since one sentence can refer to multiple slots, the problem becomes a multi-label classification problem \cite{tsoumakas2006multi}. The training data is generated by using several template sentences based on regular expressions, with different times of the day and with many linguistic variations, along with the target slot mappings. Each word of the sequence is represented as a one hot encoding and the sequence of one hot word vectors is fed directly to the LSTM. A depiction of the network utilized for solving the multi-label classification problem that utilizes LSTM units and sigmoidal output units as shown in figure \ref{fig:slotmap}. We can have one single loss function computed over all 40 output units or 40 separate loss functions, each over a single timeslot output. Results show that having separate loss functions over individual outputs yield superior performance, which appears to be in line with recent findings in the deep multi-task learning \cite{Ruder17a} literature i.e. treating each slot prediction as a seperate task allows for richer embeddings.

\begin{center}
\begin{table}
\caption{Output having same error function}
\begin{tabular}{ |c|c|c|c|c|c| } 
\hline
Model & Activation & Precision & Recall & F1-score \\
\hline
LSTM & \shortstack{Sigmoid} & 95.9\% & 94.8\% & 95.3\%\\
biLSTM & \shortstack{Sigmoid} & 98.3\% & 97.2\% & 97.5\%\\
\hline
\end{tabular}
\end{table}
\label{tab:common}
\end{center}

\begin{center}
\begin{table}
\caption{Output having separate error function}
\begin{tabular}{ |c|c|c|c|c|c| } 
\hline
Model & Activation & Precision & Recall & F1-score \\
\hline
LSTM & \shortstack{Softmax} & 97.2\% & 97.8\% & 97.3\%\\
biLSTM & \shortstack{Softmax} & 99.8\% & 99.4\% & 99.6\%\\
LSTM & \shortstack{Sigmoid} & 98.3\% & 98.3\% & 98.3\%\\
biLSTM & \shortstack{Sigmoid} & 99.9\% & 99.8\% & 99.8\%\\
\hline
\end{tabular}
\end{table}
\label{tab:separate}
\end{center}

The model has an LSTM or Bi-LSTM layer, followed by a layer of fully connected ReLU units (after tuning over several architectures), we used the Adam optimizer \cite{KingmaB14}, a mean squared error loss function and tried both sigmoidal or softmax units at the output (softmax units could only be used when we computed error individually over each output timeslot). While using sigmoid activation function, any value greater than 0.5 is considered as 1 and less than 0.5 is considered as 0. The dataset is created with different possible template phrases for conveying time information in english before using these to train the model. A dataset consisting of 1056 samples. This was divided into training, validation and testing with a ratio 0.6:0.2:0.2 \footnote{The dataset is available at \url{https://github.com/vishwa15/timephrase_data.git}.}. 

Some sample phrases used to generate the data: \textit{early morning, morning, late morning, afternoon, early afternoon, late afternoon, after lunch, before lunch, evening, early evening, late evening}

\subsection{Reinforcement Learning for Scheduling}
We design a policy gradient based reinforcement learning system that interacts with the different users to determine an agreeable slot for all the participants. The state space of the RL system encodes 1) the participants designation as a one hot encoding, 2) the current occupancy of the slots, 3) a waiting queue of the next seven meetings and the duration (number of slots) of each meeting waiting in the queue. The environment continuously introduces 1,2,4 and 6 slot meetings into the backlog and as  meetings get scheduled meetings are popped from the backlog into the waiting queue at every timestep.

\begin{figure}[h]
\includegraphics[width=\columnwidth]{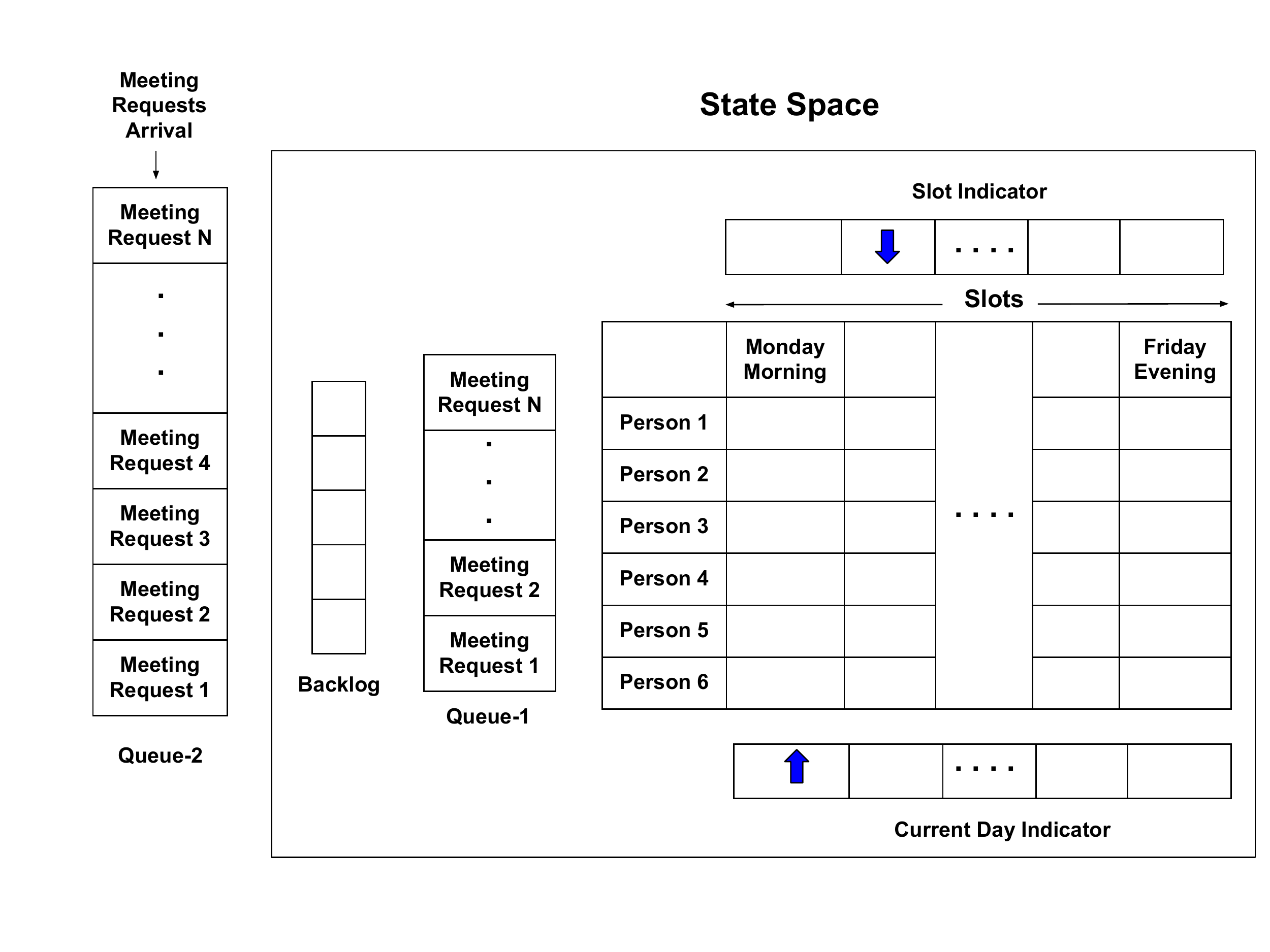}
\caption{Skeletal diagram of the meeting scheduling system \label{Model:arch}}
\end{figure}

\subsection{Formulation}
\begin{figure}[h]
\includegraphics[width=7cm, height=4cm]{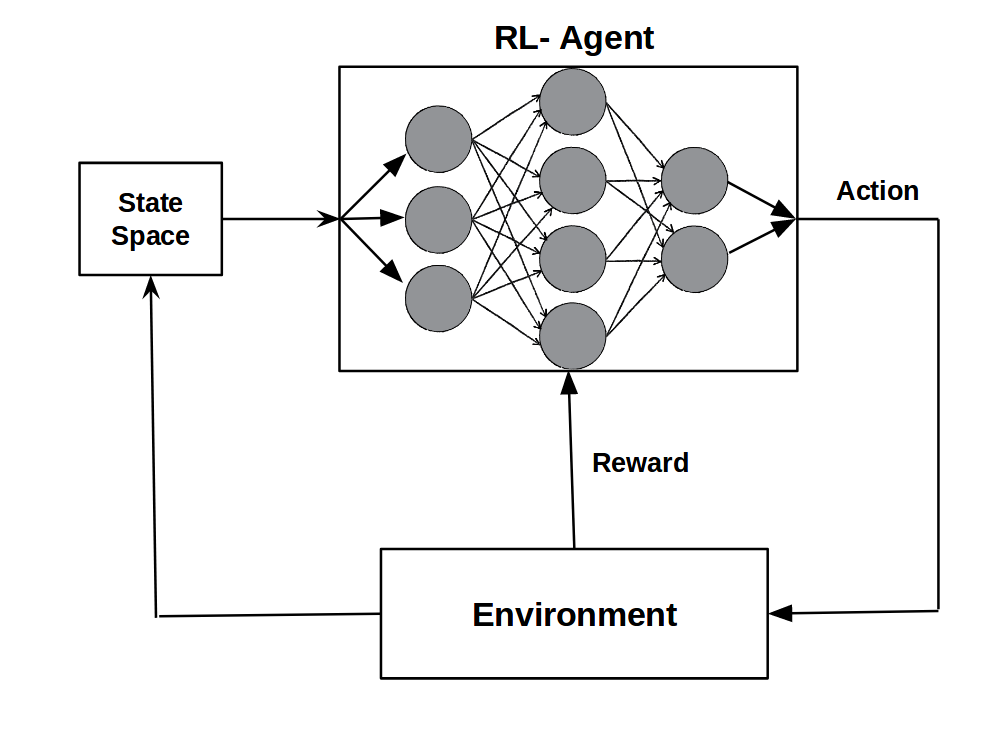}
\caption{Reinforcement Learning with Neural Networks}
\label{fig:rl_basic}
\end{figure}

As shown in figure 3, the meeting scheduling system consists of 40 slots which can be mapped to a week (5 days x 8 slots), two row matrices to indicate the slot to schedule a meeting and current day of the week. Initially, the meetings are sent into a backlog queue as they arrive and then they are let into waiting queue. The entire system except the backlog queue is given as input state to the RL agent.

Initially, the waiting queue will be filled with meeting requests. The environment selects a slot so that the meeting at the first entry of the waiting queue is schedulable and the slot indicator is placed at that position. Then the RL-agent makes a binary decision about whether or not to schedule the meeting at that position. Accordingly,  the meeting will either be requested for all the participants or will be pushed back to backlog. Once the decision is made on all the meetings in the waiting queue, the current day indicator is moved by 8 slots in a circular fashion and the 8 slots are freed up for any meetings still waiting  to get scheduled. Basically, the system schedules meetings for the next 5 days of the week and clears the slots of the past day. The fresh meetings then arrive into the waiting queue and scheduling them is treated as one step for the learning environment.

\begin{algorithm}
\For{each episode}
{
	\For{each timestep}
	{
	record $(s_1^i, a_1^i) (s_2^i, a_2^i) . . . (s_n^i, a_n^i)$\\
	compute benchmark \& meetings\_scheduled\\
	
	\uIf{meetings\_scheduled $>$= benchmark}
	{
	reward:= +1 $\forall$ $(s^i,a^i)$
	}
	\Else
	{
	reward:= -1 $\forall$ $(s^i,a^i)$
	}
	
	train using $(s_1^i, a_1^i, r_1^i) (s_2^i, a_2^i, r_2^i) . . . (s_n^i, a_n^i, r_n^i)$
	}
}
\caption{Training the RL-Agent}
\end{algorithm}

\subsection{Persistent Exploration}
As with most reinforcement learning approaches, a balance has to be maintained between exploration of new trajectories and exploitation of the knowledge of the state action space learned so far. In our formulation we allow for exploration by allowing the agent to choose a random action with a probability of 0.1 at every step. In general after training an RL-agent, the policy is usually fixed and the agent is then deployed with the learnt policy. However, in our application, the agent is allowed to continuously learn and explore with a constant probability. The reason for this is that the environment may change in unforseeable ways and the agent must be able to adapt continuously. The motivations for this  will become clearer in the experiments section. 

\section{Experiments and Results}
The performance of the RL-agent is evaluated based on the number of meetings scheduled per episode, as compared to the optimal policy (which in this case corresponds to shortest job first), a first come first serve policy, and a random policy which schedules meetings in a random order. A meeting can request for 1, 2, 4, or 6 slots and these are generated with a probability (0.4, 0.2, 0.2, 0.2) respectively (this is not necessary but these probabilities allow us to see some interesting behaviours of the RL agent). The meeting requests are let into the waiting queue from backlog based on the current arrival rate. The state space is fed as input to a neural network which has two hidden layers of length 128x32. The degree of fullness of the backlog queue is represented with a backlog vector of size 5. We utilize binary crossentropy as our loss function, Adam as our optimizer and a softmax output activation function. Note that learning is taking place continuously as mentioned in the previous Section. Also, the RL agent with same neural architecture can optimise itself to perform different tasks for a given reward function and that the state space remains the same while achieving different objectives. 
\subsection{Objective 1: Scheduling maximum no. of meetings with delayed reward}
A benchmark is calculated at the begining of each timestep which calculates the ratio of free slots to average slots per meeting. The RL-agent makes a decision on all the meetings in the waiting queue and the (state, action) pair is recorded. The number of meetings scheduled by the RL-agent in that timestep is noted. If this number is greater than or equal to the benchmark calculated then all the (state, action) pairs are given a +1 reward else the pairs receive a -1 reward. The successful experiences/timesteps which were rewarded positively are stored in a replay buffer of size 20 and used for training the  RL-agent at the end of each episode. The old experiences which are more than 20 episodes old are popped out of the replay buffer to make sure the RL-agent is learning new experiences. The number of meetings scheduled by the RL agent over multiple timesteps is recorded  and the average is calculated. With the same conditions, the average number of meetings scheduled using different policies are calculated for comparision. 	

\begin{align*}
benchmark = {\frac {total\  no.\ of\ free\ slots\ available}{\frac {slots\ required\ for\ meetings\ in\ Q1}{no.\ of\ meetings\ in\ Q1}}}
\end{align*}

\begin{figure}[h]
\includegraphics[width=\columnwidth]{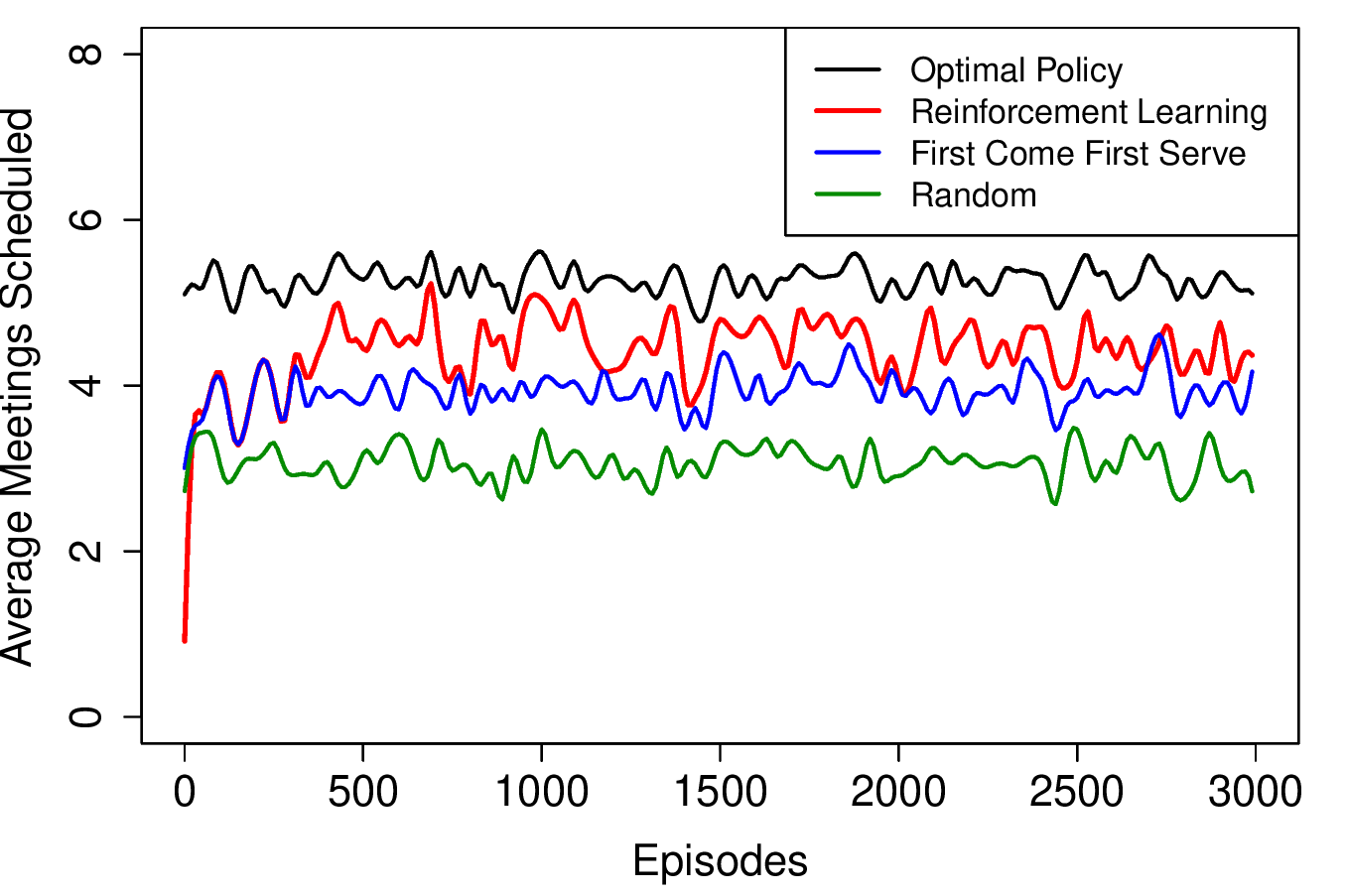}
\caption{Average no. of meetings scheduled v/s Episodes when the load is 190 to 210\% of the scheduling capacity}
\label{fig:190capacity}
\end{figure}

As shown in the figure ~\ref{fig:190capacity} initially the RL-agent schedules the meetings as they come and the number of meetings scheduled is less than the benchmark calculated. So the agent receives a -ve reward and modifies its action to learn a better policy to accommodate more meetings in the available slots. Its clear from the figure that the RL-agent tries to learn the optimal policy in order to get the maximum positive rewards. 

\begin{figure}[h]
\includegraphics[width=\columnwidth]{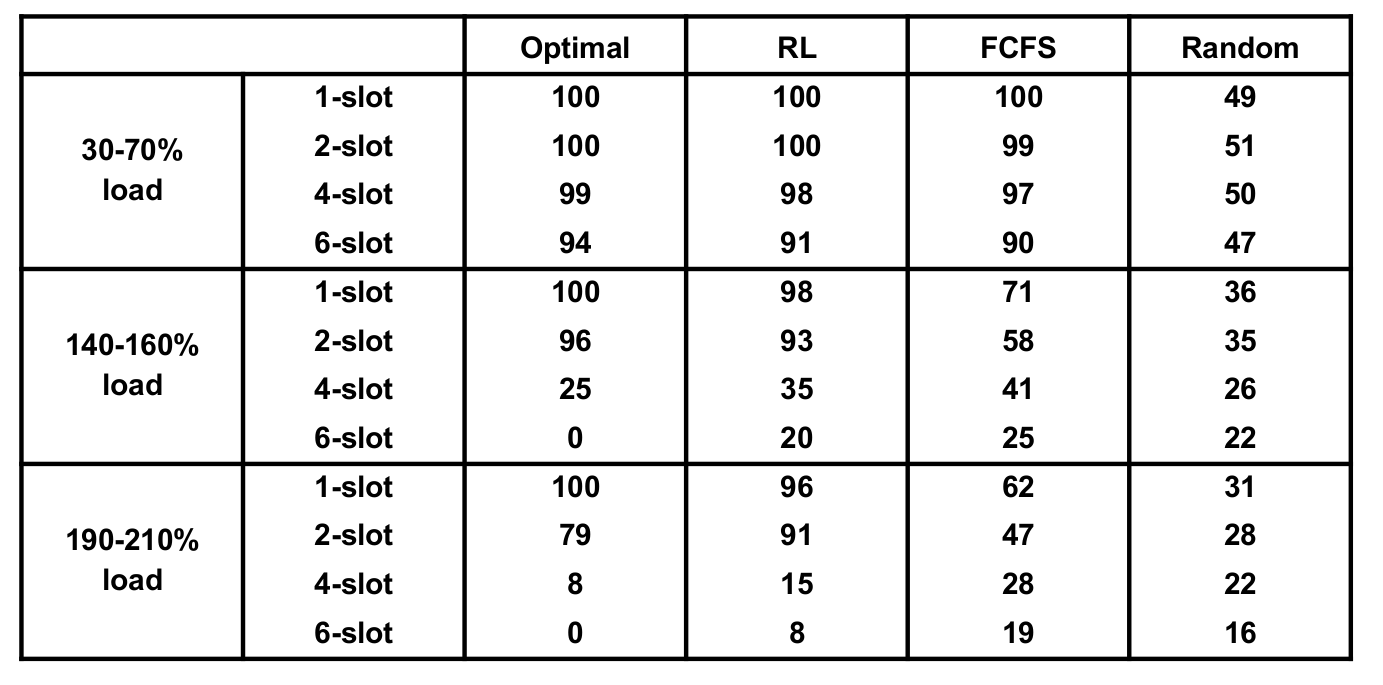}
\caption{Percentage no. of meetings scheduled by different policies at different loads}
\label{fig:table}
\end{figure}

When the available slots are fewer than those required, the RL-agent pushes the heavy (4 and 6 slot) meetings into the backlog and all the heavy meetings end up getting scheduled at the end. The episode is run until all the meetings are scheduled but for measuring the performance of the model, the timesteps which have new meetings are considered. 

\begin{figure}[h]
\includegraphics[width=\columnwidth]{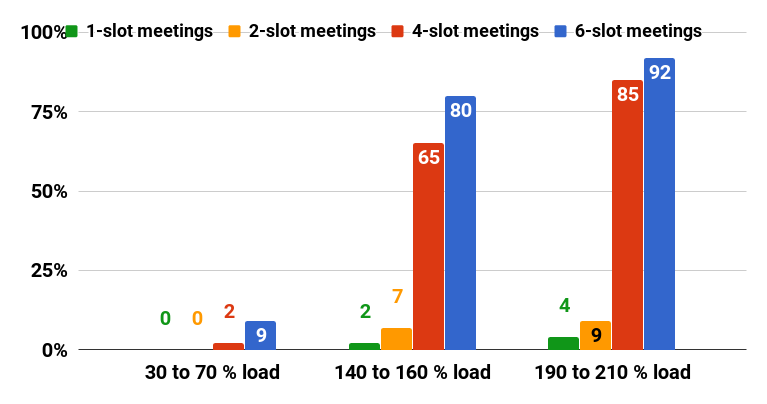}
\caption{Percentage no. of times different types of meetings are being pushed back into backlog v/s different load}
\label{fig:differentload}
\end{figure}

The same experiment is conducted with different meeting arrival loads. As shown in the Figure ~\ref{fig:differentload}, when the load is just 30 to 70\% of the scheduling capacity, the number of meetings that get rejected are very low. When the load is increased to 140-160\%  there is a sharp increase in the number of 4-slot and 6-slot meetings getting rejected and this increases further when the load is increased to 190 - 210\% of the scheduling capacity.

\subsection{Objective 2: Varying the load of the meeting arrival and learn different policy according to the environment}

\begin{figure}[h]
\includegraphics[width=\columnwidth]{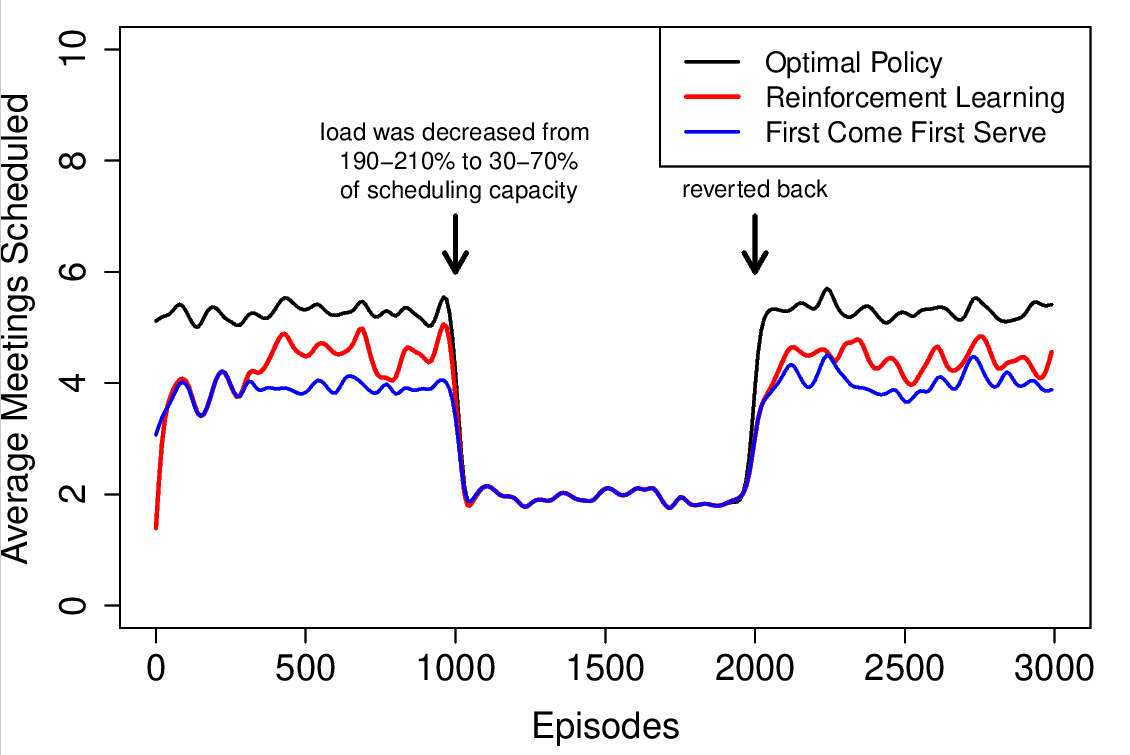}
\caption{Average no. of meetings scheduled v/s Episodes when the load was changing}
\label{fig:varyingload}
\end{figure}

The Reinforcement learning formulation and reward function is kept constant but the meeting arrival rate is suddenly changed. As shown in the Figure ~\ref{fig:varyingload}, the load of meeting arrival was kept at 190 - 210\% of the scheduling capacity at the beginning and at episode-1000 the arrival rate was decreased to 30 - 70 \%. Soon within a couple of episodes, the RL-agent learns to accept all types of meeting requests since the load is lighter. Once again at episode-2000, the arrival rate was increased and the RL-agent takes a couple of more episodes to learn a better policy to suit the environment.

\begin{figure}[h]
\includegraphics[width=\columnwidth]{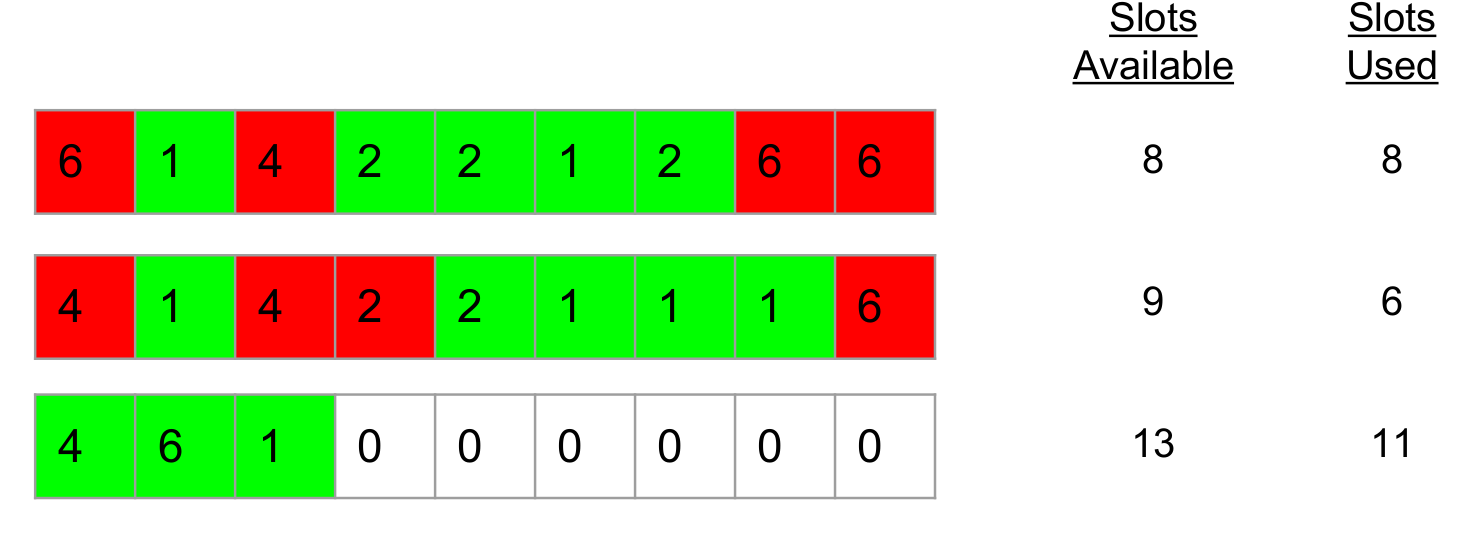}
\caption{Typical cases of meeting arrival \& their schedulings. Green indicates acceptance \& Red indicate rejection}
\label{fig:typical_case}
\end{figure}

\subsection{Objective 3: Avoid uncomfortable slots and adapt to changing preference with immediate reward}
\begin{figure}[h]
\includegraphics[width=\columnwidth]{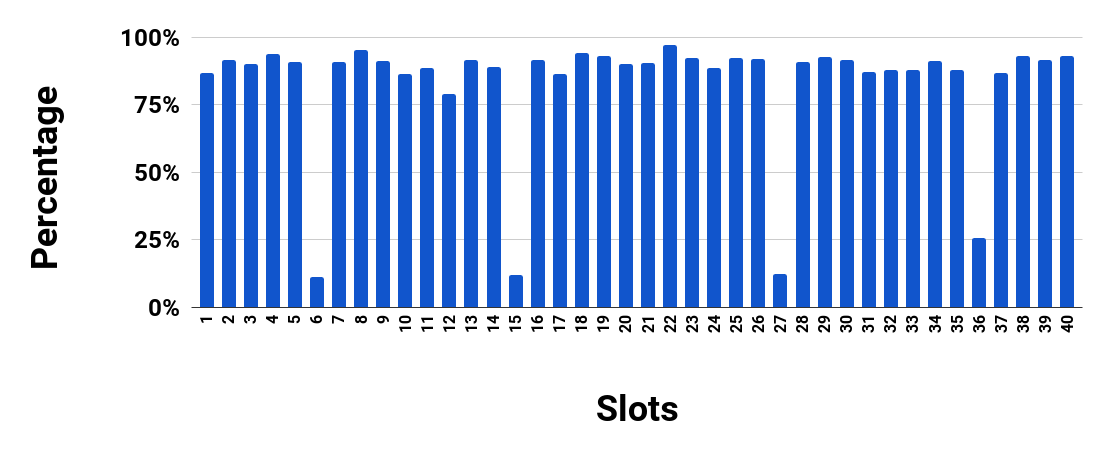}
\caption{Average no. of asks for all slots when slots 5, 14, 26, 35 was considered uncomfortable by people}
\label{fig:uncomfortableslots}
\end{figure}

\begin{figure}[h]
\includegraphics[width=\columnwidth]{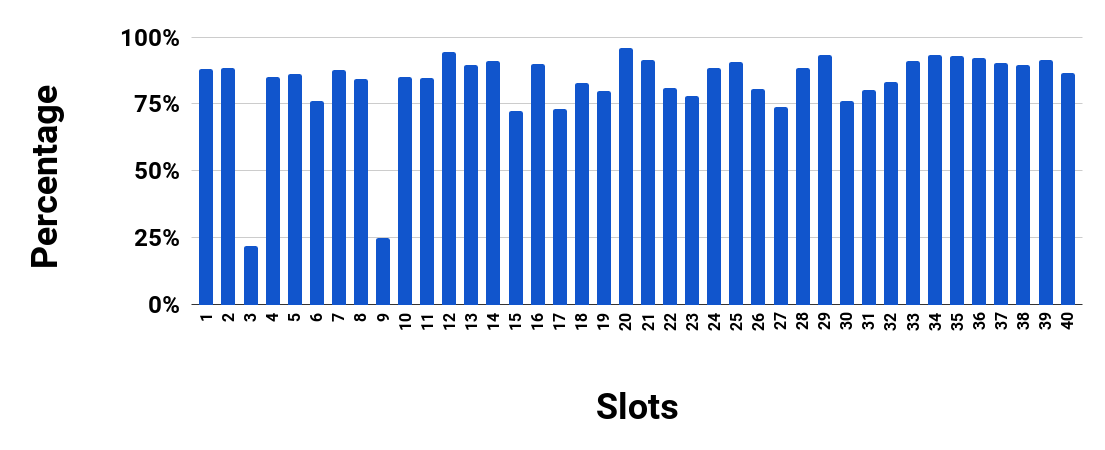}
\caption{Average no. of asks for all slots when uncomfortability was changed from slots 5, 14, 26, 35  to slots 2, 9}
\label{fig:uncomfortableslots_changed}
\end{figure}

In the experiments so far, the RL-agent picks a vacant slot where a meeting can be scheduled and the participants will be requested for that slot. However, all the participants need not necessarily agree on a meeting slot if they are busy or the slot is otherwise inconvenient (Monday mornings for example may be busy). In this experiment, we will program the environment to make a few slots uncomfortable for the users and when the RL-agent requests a meeting in those slots, the participants refuse. In this case, an immediate negative reward (-1) is given to the agent for asking to schedule a meeting at an unsuitable slot. Conversely, when a participant accepts the meeting request, the agent receives an immediate positive reward (+1). Initially slots -6, 15, 27, 36 were made uncomfortable and the RL-agent was given an immediate reward whenever these slots were requested. As shown in Figure ~\ref{fig:uncomfortableslots}, the no. of requests that were made on those uncomfortable slots reduces drastically as the agent adapts and learns to avoid them. Within the same experimental setup, we change the uncomfortable slots from slots 6, 15, 27, 36 to slots 3, 9. We can see in Figure ~\ref{fig:uncomfortableslots_changed}, the RL-agent can adapt to this changing environment

\subsection{Objective 4: Adapt to changing preference when a senior designated person asks for a meeting}
\begin{figure}[h]
\includegraphics[width=\columnwidth]{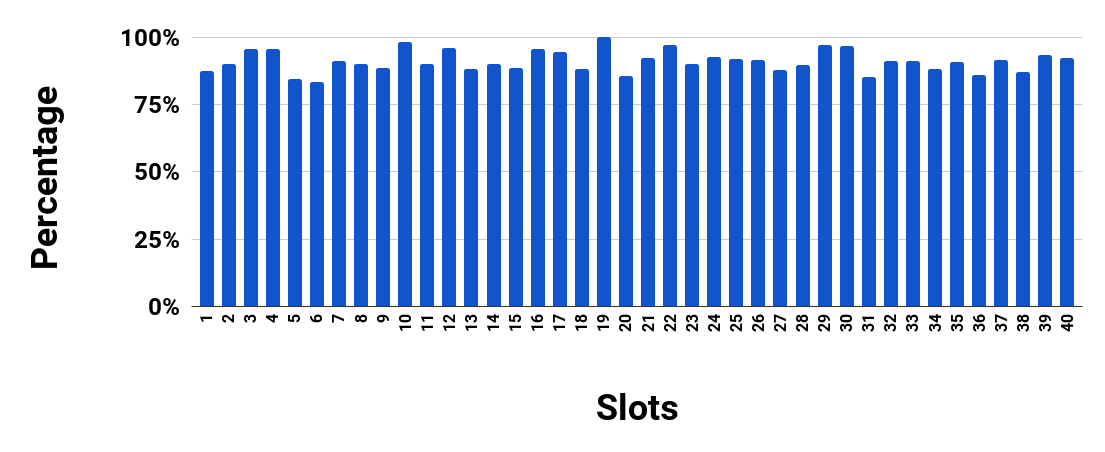}
\caption{Average no. of asks for all slots when a person with higher designation requests for a meeting}
\label{fig:senior_designation}
\end{figure}

Participants may have their own preference for slots but when a person with a senior designation asks for a meeting,  participants usually agree regardless. This behavior was implemented in the environment and the RL-agent is able to adapt to this behavior. Meeting requests consist of three fields: participants, initiator ID and slot type. The RL-agent will pick up a signal from the initiator ID and when a senior designated person requests for a meeting slot, the agent goes ahead and requests all the participants for the slot since the environment is programmed for them to agree to that slot if the initiator has a senior designation. As shown in Figure ~\ref{fig:senior_designation}, slots 5, 14, 26, 35 were made uncomfortable for the participants and yet the RL-agent utilizes these slots when a senior designated person requests a meeting. 

\section{Conclusion}
In this paper, we have designed the architecture of a meeting bot which can schedule meetings through dialogue. A multi-label classification model to convert english phrases which have time information into slots was employed with seperate output loss functions for each time slot. Due to time constraints, scheduling the maximum number of meetings is important and a model using reinforcement learning is trained to schedule them efficiently. The model can adapt to new situations with varying meeting arrival rates and the performance of the model is compared with  standard schedulers. We have also shown that the RL-agent can adapt to user preferences and schedule meetings accordingly and can also change its policy when the meeting is called by members who are at a more senior designation. This adaptive behavior cannot be replicated via a fixed scheduling policy like first come first serve or shortest job first.

\bibliographystyle{aaai}
\bibliography{tmp}

\end{document}